\documentclass{sig-alternate-05-2015}

\usepackage[T1]{fontenc}
\usepackage{subfig}
\usepackage{bbold}
\usepackage[scaled=.92]{helvet} 
\usepackage{graphicx} 
\usepackage{balance}  
\usepackage{booktabs} 
\usepackage{ccicons}  
\usepackage{ragged2e} 
\usepackage{bbm}
\usepackage[font=bf]{caption}
\usepackage[disable]{todonotes}
\usepackage{hyperref}
\usepackage{algorithm}
\usepackage{algpseudocode}
\usepackage{microtype}
\usepackage[numbers]{natbib}
\usepackage{enumitem}
\DeclareMathOperator*{\argmin}{argmin}
\DeclareMathOperator*{\argmax}{argmax}
\newcommand{\norm}[1]{\left\lVert#1\right\rVert}
\def\Sim{\pi}
\def\EModel{\xi}
\newcommand{\sameer}[1]{\todo[inline,color=green!20]{#1}}
\newcommand{\marco}[1]{\todo[inline,color=blue!20]{#1}}
\def\shortName{LIME}

\def\instanceBudget{B}

\newcommand{\presec}{\vspace{-.11in}}

\newcommand{\precap}{\vspace{-.05in}}
\newcommand{\postcap}{\vspace{-.21in}}
\newcommand{\postcapsmall}{\vspace{-.11in}}

\begin{document}


\setcopyright{acmlicensed}




\CopyrightYear{2016} 
\conferenceinfo{KDD '16,}{August 13 - 17, 2016, San Francisco, CA, USA}
\isbn{978-1-4503-4232-2/16/08}\acmPrice{\$15.00}
\doi{http://dx.doi.org/10.1145/2939672.2939778}

\clubpenalty=10000 
\widowpenalty = 10000

%
\conferenceinfo{KDD}{2016 San Francisco, CA, USA}

\title{``Why Should I Trust You?''\\Explaining the Predictions of Any Classifier}
%
%
%
%
%

\numberofauthors{3} 
%
\author{
%
%
\alignauthor
    Marco Tulio Ribeiro\\
    \affaddr{University of Washington} \\
    \affaddr{Seattle, WA 98105, USA} \\
    \email{marcotcr@cs.uw.edu}
\alignauthor
Sameer Singh\\
    \affaddr{University of Washington}\\
    \affaddr{Seattle, WA 98105, USA}\\
    \email{sameer@cs.uw.edu}
\alignauthor
Carlos Guestrin\\
    \affaddr{University of Washington}\\
    \affaddr{Seattle, WA 98105, USA}\\
    \email{guestrin@cs.uw.edu}
}
\date{\today}

\maketitle
\begin{abstract}

Despite widespread adoption, machine learning models remain mostly black boxes.
Understanding the reasons behind predictions is, however, quite important in assessing \emph{trust}, which is fundamental if one plans to take action based on a prediction, or when choosing whether to deploy a new model.
Such understanding also provides insights into the model, which can be used to transform an untrustworthy model or prediction into a trustworthy one. 

In this work, we propose \shortName{}, a novel explanation technique that explains the predictions of \emph{any} classifier in an interpretable and faithful manner, by learning an interpretable model locally around the prediction.
We also propose a method to explain models by presenting representative individual predictions and their explanations in a non-redundant way, framing the task as a submodular optimization problem.
We demonstrate the flexibility of these methods by explaining different models for text (e.g. random forests) and image classification (e.g. neural networks).
We show the utility of explanations via novel experiments, both simulated and with human subjects, on
 various scenarios that require trust: deciding if one should trust a prediction, choosing between models, improving an untrustworthy classifier, and identifying why a classifier should not be trusted.
\end{abstract}


\section{Introduction}
\noindent
Machine learning is at the core of many recent advances in science and technology.
Unfortunately, the important role of humans is an oft-overlooked aspect in the field.
Whether humans are directly using machine learning classifiers as tools, or are deploying models within other products, a vital concern remains: \emph{if the users do not trust a model or a prediction, they will not use it}. 
It is important to differentiate between two different (but related) definitions of trust: (1)~\emph{trusting a prediction}, i.e. whether a user trusts an individual prediction sufficiently to take some action based on it, and
(2)~\emph{trusting a model}, i.e. whether the user trusts a model to behave in reasonable ways if deployed.
Both are directly impacted by how much the human understands a model's behaviour, as opposed to seeing it as a black box.

Determining trust in individual predictions is an important problem when the model is used for decision making. 
When using machine learning for medical diagnosis~\cite{caruana2015} or terrorism detection, for example, predictions cannot be acted upon on blind faith, as the consequences may be catastrophic. 

Apart from trusting individual predictions, there is also a need to evaluate the model as a whole before deploying it ``in the wild''.
To make this decision, users need to be confident that the model will perform well on real-world data, according to the metrics of interest. 
Currently, models are evaluated using accuracy metrics on an available validation dataset.
However, real-world data is often significantly different, and further, the evaluation metric may not be indicative of the product's goal. 
Inspecting individual predictions and their explanations is a worthwhile solution, in addition to such metrics. 
In this case, it is important to aid users by suggesting which instances to inspect, especially for large datasets.

In this paper, we propose providing explanations for individual predictions as a solution to the ``trusting a prediction'' problem, and selecting multiple such predictions (and explanations) as a solution to the ``trusting the model'' problem.
Our main contributions are summarized as follows.
\begin{itemize}[itemsep=0.5ex,partopsep=1ex,parsep=1ex,leftmargin=*] 
\item \shortName{}, an algorithm that can explain the predictions of \emph{any} classifier or regressor in a faithful way, by approximating it locally with an interpretable model.
\item SP-\shortName{}, a method that selects a set of representative instances with explanations to address the ``trusting the model'' problem, via submodular optimization.
\item Comprehensive evaluation with simulated and human subjects, where we measure the impact of explanations on trust and associated tasks. 
In our experiments, non-experts using \shortName{} are able to pick which classifier from a pair generalizes better in the real world.
Further, they are able to greatly improve an untrustworthy classifier trained on 20 newsgroups, by doing feature engineering using \shortName{}.
We also show how understanding the predictions of a neural network on images helps practitioners know when and why they should not trust a model.
\end{itemize}

\begin{figure*}
    \centering
    \includegraphics[width=0.75\textwidth]{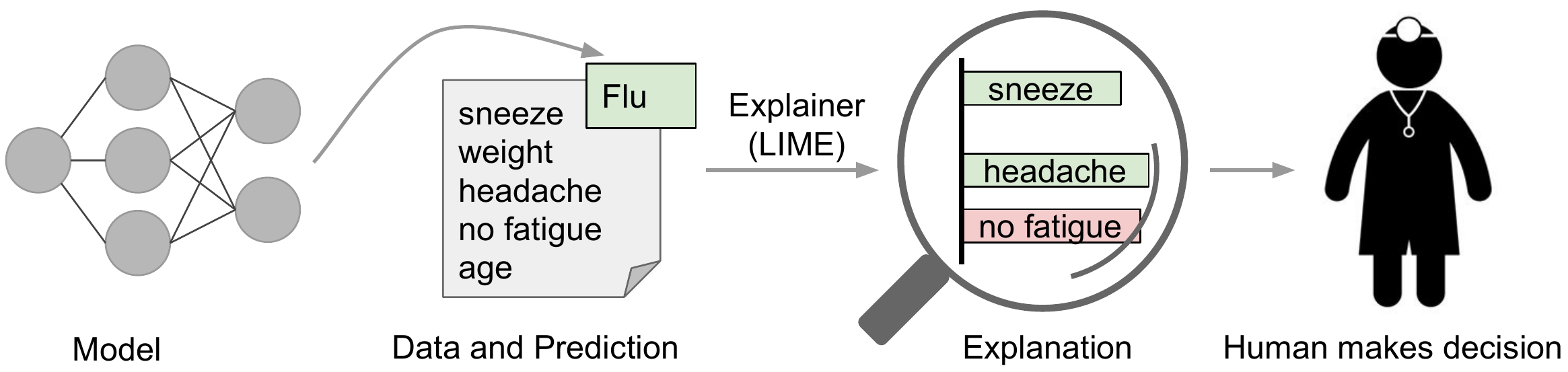}
    \precap
    \caption{Explaining individual predictions. A model predicts that a patient
    has the flu, and \shortName{} highlights the symptoms in the patient's
    history that led to the prediction. Sneeze and headache are portrayed as
    contributing to the ``flu'' prediction, while ``no fatigue'' is evidence
    against it. With these, a doctor can make an informed decision about whether to trust the model's prediction.}~\label{fig:trust_flowchart1}
    \postcap
\end{figure*}

\section{The case for explanations}
\label{sec:explain}
By ``explaining a prediction'', we mean presenting textual or visual artifacts that provide qualitative understanding of the relationship between the instance's components (e.g. words in text, patches in an image) and the model's prediction.
We argue that explaining predictions is an important aspect in getting humans to trust and use machine learning effectively, if the explanations are faithful and intelligible.

The process of explaining individual predictions is illustrated in Figure~\ref{fig:trust_flowchart1}. 
It is clear that a doctor is much better positioned to make a decision with the help of a model if intelligible explanations are provided.
In this case, an explanation is a small list of symptoms with relative weights -- symptoms that either contribute to the prediction (in green) or are evidence against it (in red). 
Humans usually have prior knowledge about the application domain, which they can use to accept (trust) or reject a prediction if they understand the reasoning behind it.
It has been observed, for example, that providing explanations can increase the acceptance of movie recommendations~\cite{recsys} and other automated systems~\cite{dzindolet}. 

Every machine learning application also requires a certain measure of overall trust in the model.
Development and evaluation of a classification model often consists of collecting annotated data, of which a held-out subset is used for automated evaluation.  
Although this is a useful pipeline for many applications, evaluation on validation data may not correspond to performance ``in the wild'', as practitioners often overestimate the accuracy of their models~\cite{Patel:2008:ISM:1357054.1357160}, and thus trust cannot rely solely on it.
Looking at examples offers an alternative method to assess truth in the model, especially if the examples are explained. 
We thus propose explaining several representative individual predictions of a model as a way to provide a global understanding. 

There are several ways a model or its evaluation can go wrong.
Data leakage, for example, defined as the unintentional leakage of signal into the training (and validation) data that would not appear when deployed~\cite{leakage}, potentially increases accuracy.
A challenging example cited by \citet{leakage} is one where the patient ID was found to be heavily correlated with the target class in the training and validation data.
This issue would be incredibly challenging to identify just by observing the predictions and the raw data, but much easier if explanations such as the one in Figure~\ref{fig:trust_flowchart1} are provided, as patient ID would be listed as an explanation for predictions.
Another particularly hard to detect problem is dataset shift \cite{datashift}, where training data is different than test data (we give an example in the famous 20 newsgroups dataset later on).
The insights given by explanations 
are particularly helpful in identifying what must be done to convert an untrustworthy model into a trustworthy one -- for example, removing leaked data or changing the training data to avoid dataset shift.

\begin{figure}[tb]
    \centering
    \includegraphics[width=\columnwidth]{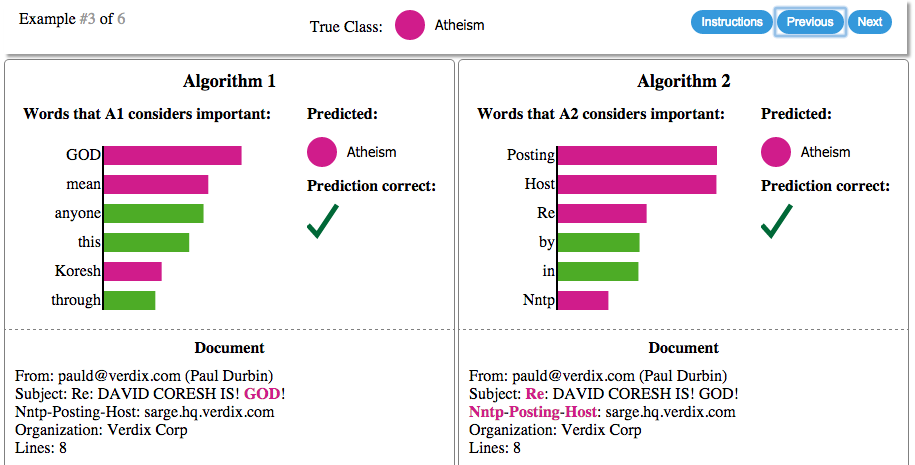}
    \precap
    \precap
    \precap
    \caption{Explaining individual predictions of competing classifiers trying
    to determine if a document is about ``Christianity'' or ``Atheism''.
    The bar chart represents the importance given to the most relevant words,
    also highlighted in the text.
    Color indicates which class the word contributes to (green for ``Christianity'', magenta for ``Atheism'').
    }~\label{fig:mturkinterface1}
    \postcap
    \vspace{-0.05in}
\end{figure}

Machine learning practitioners often have to select a model from a number of alternatives, requiring them to assess the relative trust between two or more models.
In Figure \ref{fig:mturkinterface1}, we show how individual prediction explanations can be used to select between models, in conjunction with accuracy.
In this case, the algorithm with higher accuracy on the validation set is actually much worse, a fact that is easy to see when explanations are provided (again, due to human prior knowledge), but hard otherwise.
Further, there is frequently a mismatch between the metrics that we can compute and optimize (e.g. accuracy) and the actual metrics of interest such as user engagement and retention.
While we may not be able to measure such metrics, we have knowledge about how certain model behaviors can influence them. 
Therefore, a practitioner may wish to choose a less accurate model for content recommendation that does not place high importance in features related to ``clickbait'' articles (which may hurt user retention), even if exploiting such features increases the accuracy of the model in cross validation.
We note that explanations are particularly useful in these (and other) scenarios if a method can produce them for \emph{any} model, so that a variety of models can be compared.

\subsubsection*{Desired Characteristics for Explainers}
We now outline a number of desired characteristics from explanation methods. 

An essential criterion for explanations is that they must be \textbf{interpretable}, i.e., provide qualitative understanding between the input variables and the response. 
We note that interpretability must take into account the user's limitations.
Thus, a linear model~\cite{gametheory}, a gradient vector~\cite{Baehrens:2010:EIC:1756006.1859912} or an additive model~\cite{caruana2015} may or may not be interpretable.
For example, if hundreds or thousands of features significantly contribute to a prediction, it is not reasonable to expect any user to comprehend why the prediction was made, even if individual weights can be inspected. 
This requirement further implies that explanations should be easy to understand, which is not necessarily true of the features used by the model, and thus the ``input variables'' in the explanations may need to be different than the features.
Finally, we note that the notion of interpretability also depends on the target audience.
Machine learning practitioners may be able to interpret small Bayesian networks, but laymen may be more comfortable with a small number of weighted features as an explanation.

Another essential criterion is \textbf{local fidelity}.
Although it is often impossible for an explanation to be completely faithful unless it is the complete description of the model itself, for an explanation to be meaningful it must at least be \emph{locally faithful}, i.e. it must correspond to how the model behaves in the vicinity of the instance being predicted.
We note that local fidelity does not imply global fidelity: features that are globally important may not be important in the local context, and vice versa.
While global fidelity would imply local fidelity, identifying globally faithful explanations that are interpretable remains a challenge for complex models.

While there are models that are inherently interpretable~\cite{caruana2015,LethamRuMcMa15,supersparse,WangRu15}, an explainer should be able to explain \emph{any} model, and thus be \textbf{model-agnostic} (i.e. treat the original model as a black box).
Apart from the fact that many state-of-the-art classifiers are not currently interpretable, this also provides flexibility to explain future classifiers.

In addition to explaining predictions, providing a \textbf{global perspective} is important to ascertain trust in the model.
As mentioned before, accuracy may often not be a suitable metric to evaluate the model, and thus we want to \emph{explain the model}.
Building upon the explanations for individual predictions, we select a few explanations to present to the user, such that they are representative of the model.
\presec
\section{Local Interpretable \\Model-Agnostic Explanations}
\label{sec:lime}

We now present Local Interpretable Model-agnostic Explanations (\textbf{\shortName{}}).
The overall goal of \shortName{} is to identify an \textbf{interpretable} model over the \emph{interpretable representation} that is \textbf{locally faithful} to the classifier.

\subsection{Interpretable Data Representations}
Before we present the explanation system, it is important to distinguish between features and interpretable data representations.
As mentioned before, \textbf{interpretable} explanations need to use a representation that is understandable to humans, regardless of the actual features used by the model.
For example, a possible \emph{interpretable representation} for text classification is a binary vector indicating the presence or absence of a word, even though the classifier may use more complex (and incomprehensible) features such as word embeddings. 
Likewise for image classification, an \emph{interpretable representation} may be a binary vector indicating the ``presence'' or ``absence'' of a contiguous patch of similar pixels (a super-pixel), while the classifier may represent the image as a tensor with three color channels per pixel.
We denote $x \in \mathbb{R}^d$ be the original representation of an instance being explained, and we use $x' \in \{0,1\}^{d'}$ to denote a binary vector for its interpretable representation.

\subsection{Fidelity-Interpretability Trade-off}

Formally, we define an explanation as a model $g \in G$, where $G$ is a class of potentially \emph{interpretable} models, such as linear models, decision trees, or falling rule lists~\cite{WangRu15}, i.e. a model $g \in G$ can be readily presented to the user with visual or textual artifacts.
The domain of $g$ is $\{0,1\}^{d'}$, i.e. $g$ acts over absence/presence of the \emph{interpretable components}.
As not every $g \in G$ may be simple enough to be interpretable - thus we let $\Omega(g)$ be a measure of \emph{complexity} (as opposed to \emph{interpretability}) of the explanation $g\in G$.
For example, for decision trees $\Omega(g)$ may be the depth of the tree, while for linear models, $\Omega(g)$ may be the number of non-zero weights.

Let the model being explained be denoted $f: \mathbb{R}^d\rightarrow \mathbb{R}$. In classification, $f(x)$ is the probability (or a binary indicator) that $x$ belongs to a certain class\footnote{For multiple classes, we explain each class separately, thus $f(x)$ is the prediction of the relevant class.}.
We further use $\Sim_{x}(z)$ as a proximity measure between an instance $z$ to $x$, so as to define locality around $x$.
Finally, let $\mathcal{L}(f, g, \Sim_x)$ be a measure of how unfaithful $g$ is in approximating $f$ in the locality defined by $\Sim_x$. 
In order to ensure both \textbf{interpretability} and \textbf{local fidelity}, we must minimize $\mathcal{L}(f, g, \Sim_x)$ while having $\Omega(g)$ be low enough to be interpretable by humans.
The explanation produced by \textbf{LIME} is obtained by the following: 
\begin{equation}
\EModel(x) = \argmin_{g \in G}\;\;\mathcal{L}(f, g, \Sim_x) + \Omega(g)
\label{eq:lime}
\end{equation}
This formulation can be used with different explanation families $G$, fidelity functions $\mathcal{L}$, and complexity measures $\Omega$.
Here we focus on sparse linear models as explanations, and on performing the search using perturbations.

\subsection{Sampling for Local Exploration}

\begin{figure}[tb]
\centering
\includegraphics[width=0.8\columnwidth]{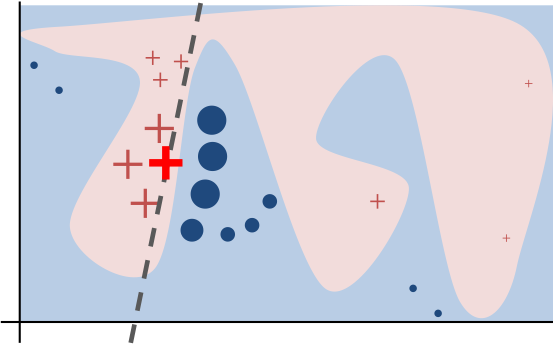}
\precap
\caption{Toy example to present intuition for LIME. The black-box model's
complex decision function $f$ (unknown to LIME) is represented by the blue/pink
background, which cannot be approximated well by a linear model. The bold red cross is the instance being explained. LIME samples instances, gets predictions using $f$, and weighs them by the proximity to the instance being explained (represented here by size). The dashed line is the learned explanation that is locally (but not globally) faithful.}
\postcap
\label{fig:lime}
\end{figure}

We want to minimize the locality-aware loss $\mathcal{L}(f, g, \Sim_x)$ without making any assumptions about $f$, since we want the explainer to be \textbf{model-agnostic}.
Thus, in order to learn the local behavior of $f$ as the interpretable inputs vary, we approximate $\mathcal{L}(f, g, \Sim_x)$ by drawing samples, weighted by $\Sim_x$.
We sample instances around $x'$ by drawing nonzero elements of $x'$ uniformly at random (where the number of such draws is also uniformly sampled).
Given a perturbed sample $z' \in \{0,1\}^{d'}$ (which contains a fraction of the nonzero elements of $x'$), we recover the sample in the original representation $z \in R^{d}$ and obtain $f(z)$, which is used as a \emph{label} for the explanation model.  
Given this dataset $\mathcal{Z}$ of perturbed samples with the associated labels, we optimize Eq.~\eqref{eq:lime} to get an explanation $\EModel(x)$.
The primary intuition behind LIME is presented in Figure~\ref{fig:lime}, where we sample instances both in the vicinity of $x$ (which have a high weight due to $\Sim_x$) and far away from $x$ (low weight from $\Sim_x$).
Even though the original model may be too complex to explain globally, LIME presents an explanation that is locally faithful (linear in this case), where the locality is captured by $\Sim_x$.
It is worth noting that our method is fairly robust to sampling noise since the samples are weighted by $\Sim_x$ in Eq.~\eqref{eq:lime}.
We now present a concrete instance of this general framework.

\subsection{Sparse Linear Explanations}

For the rest of this paper, we let $G$ be the class of linear models, such that $g(z') = w_g \cdot z'$.
We use the locally weighted square loss as $\mathcal{L}$, as defined in Eq.~\eqref{locally_weighted_loss}, where we let $\Sim_x(z) = exp(-D(x,z)^2 / \sigma^2)$ be an exponential kernel defined on some distance function $D$ (e.g. cosine distance for text, $L2$ distance for images) with width $\sigma$.
\begin{equation}
\mathcal{L}(f, g, \Sim_x) = \sum_{z, z' \in \mathcal{Z}}\Sim_x(z)\left(f(z) - g(z')\right)^2
\label{locally_weighted_loss}
\end{equation}
For text classification, we ensure that the explanation is \textbf{interpretable} by letting the \emph{interpretable representation} be a bag of words, and by setting a limit $K$ on the number of words, i.e. $\Omega(g) = \infty\mathbb{1}[\norm{w_g}_0 > K]$.
Potentially, $K$ can be adapted to be as big as the user can handle, or we could have different values of $K$ for different instances.
In this paper we use a constant value for $K$, leaving the exploration of different values to future work. 
We use the same $\Omega$ for image classification, using ``super-pixels'' (computed using any standard algorithm) instead of words, such that the interpretable representation of an image is a binary vector where $1$ indicates the original super-pixel and $0$ indicates a grayed out super-pixel.
This particular choice of $\Omega$ makes directly solving Eq.~\eqref{eq:lime} intractable, but we approximate it by first selecting $K$ features with Lasso (using the regularization path~\cite{lars}) and then learning the weights via least squares (a procedure we call K-LASSO in Algorithm \ref{alg:lime}).
Since Algorithm \ref{alg:lime} produces an explanation for an individual prediction, its complexity does not depend on the size of the dataset, but instead on time to compute $f(x)$ and on the number of samples $N$.
In practice, explaining random forests with $1000$ trees using scikit-learn~(\url{http://scikit-learn.org}) on a laptop with $N=5000$ takes under 3 seconds without any optimizations such as using gpus or parallelization. 
Explaining each prediction of the Inception network~\cite{inception} for image classification takes around 10 minutes.

Any choice of interpretable representations and $G$ will have some inherent drawbacks.
First, while the underlying model can be treated as a black-box, certain interpretable representations will not be powerful enough to explain certain behaviors.
For example, a model that predicts sepia-toned images to be \emph{retro} cannot be explained by presence of absence of super pixels.
Second, our choice of $G$ (sparse linear models) means that if the underlying model is highly non-linear even in the locality of the prediction, there may not be a faithful explanation.
However, we can estimate the faithfulness of the explanation on $\mathcal{Z}$, and present this information to the user.
This estimate of faithfulness can also be used for selecting an appropriate family of explanations from a set of multiple interpretable model classes, thus adapting to the given dataset and the classifier.
We leave such exploration for future work, as linear explanations work quite well for multiple black-box models in our experiments.

\begin{algorithm}[tb]
\begin{algorithmic}
  \Require Classifier $f$, Number of samples $N$
  \Require Instance $x$, and its interpretable version $x'$
  \Require Similarity kernel $\Sim_x$, Length of explanation $K$
  \State $\mathcal{Z} \gets \{\}$
  \For{$i \in \{1,2,3,...,N\}$}
    \State{$z_i'\gets sample\_around(x')$}
    \State $\mathcal{Z} \gets \mathcal{Z} \cup \langle z_i', f(z_i), \Sim_x(z_i) \rangle$
  \EndFor
  \State $w \gets \text{K-Lasso}(\mathcal{Z},K)$  \Comment{with $z_i'$ as features, $f(z)$ as target}
  \Return $w$
\end{algorithmic}
\caption{Sparse Linear Explanations using LIME\label{alg:lime}}
\end{algorithm}

\begin{figure*}
\centering
\subfloat[Original Image]{
\includegraphics[width=0.24\textwidth]{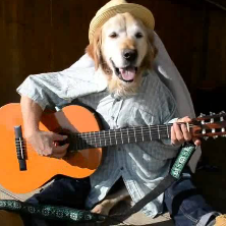}
\label{fig:original}}
\subfloat[Explaining \emph{Electric guitar}]{
\includegraphics[width=0.24\textwidth]{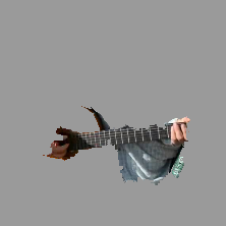}
\label{fig:electric}}
\subfloat[Explaining \emph{Acoustic guitar}]{
\includegraphics[width=0.24\textwidth]{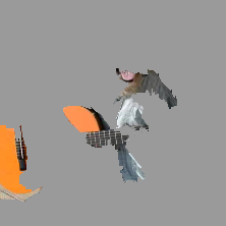}
\label{fig:acoustic}}
\subfloat[Explaining \emph{Labrador}\label{fig:labrador}]{
\includegraphics[width=0.24\textwidth]{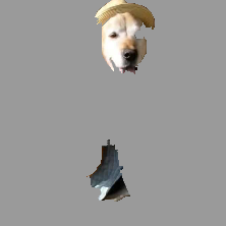}}
\precap
\caption{Explaining an image classification prediction made by Google's Inception neural network. The top 3 classes predicted are ``Electric Guitar'' ($p=0.32$), ``Acoustic guitar'' ($p=0.24$) and ``Labrador'' ($p=0.21$) }
\label{fig:inception}
\end{figure*}

\subsection{Example 1: Text classification with SVMs}
\noindent
In Figure \ref{fig:mturkinterface1} (right side), we explain the predictions of a support vector machine with RBF kernel trained on unigrams to differentiate ``Christianity'' from ``Atheism'' (on a subset of the 20 newsgroup dataset).
Although this classifier achieves $94\%$ held-out accuracy, and one would be tempted to trust it based on this, the explanation for an instance shows that predictions are made for quite arbitrary reasons (words ``Posting'', ``Host'', and ``Re'' have no connection to either Christianity or Atheism).
The word ``Posting'' appears in 22\% of examples in the training set, 99\% of them in the class ``Atheism''.
Even if headers are removed, proper names of prolific posters in the original newsgroups are selected by the classifier, which would also not generalize.

After getting such insights from explanations, it is clear that this dataset has serious issues (which are not evident just by studying the raw data or predictions), and that this classifier, or held-out evaluation, cannot be trusted.
It is also clear what the problems are, and the steps that can be taken to fix these issues and train a more trustworthy classifier.

\subsection{Example 2: Deep networks for images}
\noindent
When using sparse linear explanations for image classifiers, one may wish to just highlight the super-pixels with positive weight towards a specific class, as they give intuition as to why the model would think that class may be present.
We explain the prediction of Google's pre-trained Inception neural network \cite{inception} in this fashion on an arbitrary image (Figure~\ref{fig:original}).
Figures \ref{fig:electric}, \ref{fig:acoustic}, \ref{fig:labrador} show the superpixels explanations for the top $3$ predicted classes (with the rest of the image grayed out), having set $K=10$. 
What the neural network picks up on for each of the classes is quite natural to humans - Figure~\ref{fig:electric} in particular provides insight as to why acoustic guitar was predicted to be electric: due to the fretboard. This kind of explanation enhances trust in the classifier (even if the top predicted class is wrong), as it shows that it is not acting in an unreasonable manner.

\pagebreak
\presec
\section{Submodular Pick for\\Explaining Models}
\label{sec:submodular}

\sameer{more structure: problem, related work, our proposal, ...}

Although an explanation of a single prediction provides some understanding into the reliability of the classifier to the user, it is not sufficient to evaluate and assess trust in the model as a whole.
We propose to give a global understanding of the model by explaining a set of individual instances.
This approach is still model agnostic, and is complementary to computing summary statistics such as held-out accuracy.

Even though explanations of multiple instances can be insightful, these instances need to be selected judiciously, since users may not have the time to examine a large number of explanations.
We represent the time/patience that humans have by a budget $B$ that denotes the number of explanations they are willing to look at in order to understand a model.
Given a set of instances $X$, we define the \textbf{pick step} as the task of selecting $B$ instances for the user to inspect.

The pick step is not dependent on the existence of explanations - one of the main purpose of tools like Modeltracker~\cite{modeltracker} and others \cite{pick_kulesza} is to assist users in selecting instances themselves, and examining the raw data and predictions.
However, since looking at raw data is not enough to understand predictions and get insights, the pick step should take into account the explanations that accompany each prediction.
Moreover, this method should pick a diverse, representative set of explanations to show the user -- i.e. non-redundant explanations that represent how the model behaves globally.

\begin{figure}
\centering
\includegraphics[width=0.20\textwidth]{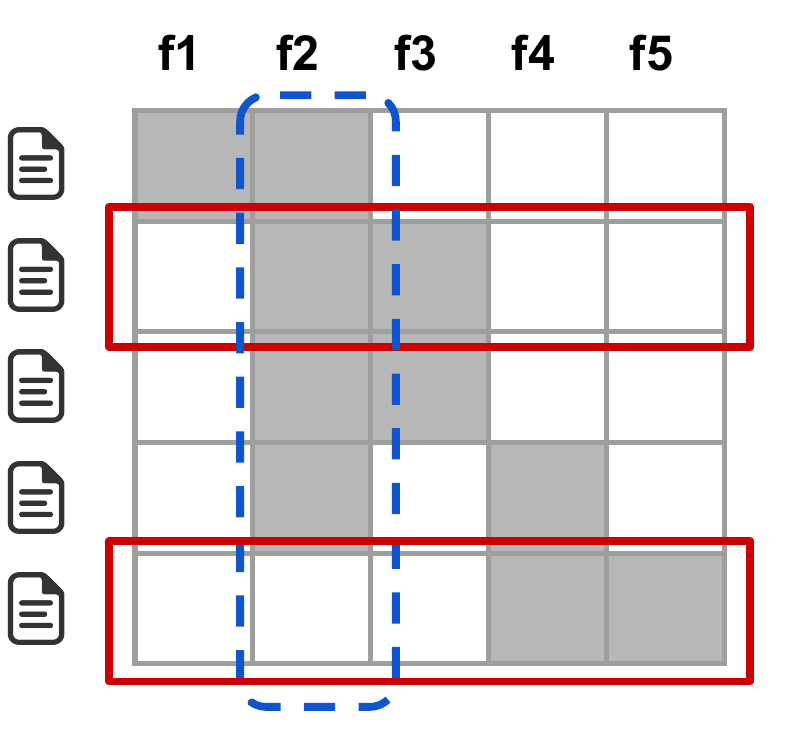}
\precap
\caption{Toy example $\mathcal{W}$. Rows represent instances (documents) and
columns represent features (words). Feature f2 (dotted
blue) has the highest importance. Rows 2 and 5 (in red) would be
selected by the pick procedure, covering all but feature f1. 
}
\label{fig:submodular}
\end{figure}

Given the explanations for a set of instances $X$ ($|X|=n$), we construct an $n\times d'$ \emph{explanation matrix} $\mathcal{W}$ that represents the local importance of the interpretable components for each instance.
When using linear models as explanations, for an instance $x_i$ and explanation $g_i=\EModel(x_i)$, we set $\mathcal{W}_{ij}=|w_{g_{ij}}|$.
Further, for each component (column) $j$ in $\mathcal{W}$, we let $I_j$ denote the \emph{global} importance of that component in the explanation space. 
Intuitively, we want $I$ such that features that explain many different instances have higher importance scores.
In Figure~\ref{fig:submodular}, we show a toy example $\mathcal{W}$, with $n=d'=5$, where $\mathcal{W}$ is binary (for simplicity). 
\marco{Rething if we want to change f1 and f2 in figure and here}
The importance function $I$ should score feature~f2 higher than feature~f1, i.e.  $I_2 > I_1$, since feature f2 is used to explain more instances.
Concretely for the text applications, we set $I_j =\sqrt{\sum_{i=1}^{n}\mathcal{W}_{ij}}$. 
For images, $I$ must measure something that is comparable across the super-pixels in different images, such as color histograms or other features of super-pixels; we leave further exploration of these ideas for future work. 

\begin{algorithm}[tb]
\begin{algorithmic}
  \Require Instances $X$, Budget $\instanceBudget$
  \ForAll{$x_i \in X$}
    \State $\mathcal{W}_i \gets \textbf{explain}(x_i, x_i')$ \Comment{Using Algorithm~\ref{alg:lime}}
  \EndFor
  \For{$j \in \{1\ldots d'\}$}
    \State $I_j \gets \sqrt{\sum_{i=1}^{n}|\mathcal{W}_{ij}|}$ \Comment{Compute feature importances}
  \EndFor
  \State $V \gets \{\}$
  \While{$|V|<\instanceBudget$} \Comment{Greedy optimization of Eq~\eqref{eq:submodular}}
    \State $V \gets V \cup \argmax_{i} c(V \cup \{i\}, \mathcal{W}, I)$
  \EndWhile
  \State \Return $V$
\end{algorithmic}
\caption{Submodular pick (SP) algorithm\label{alg:submod}
}
\end{algorithm}

While we want to pick instances that cover the important components, the set of explanations must not be redundant in the components they show the users, i.e. avoid selecting instances with similar explanations.
In Figure \ref{fig:submodular}, after the second row is picked, the third row adds no value, as the user has already seen features f2 and f3 - while the last row exposes the user to completely new features.
Selecting the second and last row results in the coverage of almost all the features.
We formalize this non-redundant coverage intuition in Eq.~\eqref{eq:coverage}, where we define coverage as the set function $c$ that, given $\mathcal{W}$ and $I$, computes the total importance of the features that appear in at least one instance in a set $V$.

\precap
\begin{equation}
c(V, \mathcal{W}, I) = \sum_{j=1}^{d'} \mathbb{1}_{[\exists i \in V : \mathcal{W}_{ij} >0]} I_j
\label{eq:coverage}
\end{equation}
The pick problem, defined in Eq.~\eqref{eq:submodular}, consists of finding the set $V, |V| \leq \instanceBudget$ that achieves highest coverage.
\begin{equation}
Pick(\mathcal{W}, I) = \argmax_{V,|V| \leq B} c(V, \mathcal{W}, I)
\label{eq:submodular}
\end{equation}
The problem in Eq.~\eqref{eq:submodular} is maximizing a weighted coverage function, and is NP-hard \cite{Feige:1998:TLN:285055.285059}.
Let $c(V \cup \{i\}, \mathcal{W}, I) - c(V, \mathcal{W}, I)$ be the marginal coverage gain of adding an instance $i$ to a set $V$.
Due to submodularity, a greedy algorithm that iteratively adds the instance with the highest marginal coverage gain to the solution offers a constant-factor approximation guarantee of  $1 - 1/e$  to the optimum~\cite{krause14survey}.
We outline this approximation in Algorithm~\ref{alg:submod}, and call it \textbf{submodular pick}.

\section{Simulated User Experiments}
\label{sec:simulated}

In this section, we present simulated user experiments to evaluate the
utility of explanations in trust-related tasks. In particular, we address the following questions:
(1)~Are the explanations faithful to the model, 
(2)~Can the explanations aid users in ascertaining trust in predictions, and
(3)~Are the explanations useful for evaluating the model as a whole.
Code and data for replicating our experiments are available at \url{https://github.com/marcotcr/lime-experiments}.

\subsection{Experiment Setup}
\label{sec:simulated:setup}
We use two sentiment analysis datasets (\emph{books} and \emph{DVDs}, 2000 instances each) where the task is to classify product reviews as positive or negative~\cite{Blitzer07Biographies}.
We train decision trees~(\textbf{DT}), logistic regression with L2 regularization~(\textbf{LR}), nearest neighbors~(\textbf{NN}), and support vector machines with RBF kernel (\textbf{SVM}), all using bag of words as features.
We also include random forests (with $1000$ trees) trained with the average word2vec embedding~\cite{wordvec} (\textbf{RF}), a model that is impossible to interpret without a technique like \shortName{}. 
We use the implementations and default parameters of scikit-learn, unless noted otherwise. 
We divide each dataset into train (1600 instances) and test (400 instances).

To explain individual predictions, we compare our proposed approach (\textbf{\shortName{}}), with \textbf{parzen}~\cite{Baehrens:2010:EIC:1756006.1859912}, a method that approximates the black box classifier globally with Parzen windows, and explains individual predictions by taking the gradient of the prediction probability function.
For parzen, we take the $K$ features with the highest absolute gradients as explanations. 
We set the hyper-parameters for parzen and \shortName{} using cross validation, and set $N=15,000$.
We also compare against a \textbf{greedy} procedure (similar to \citet{martens}) in which we greedily remove features that contribute the most to the predicted class until the prediction changes (or we reach the maximum of $K$ features), and a \textbf{random} procedure that randomly picks $K$ features as an explanation.
We set $K$ to $10$ for our experiments.

For experiments where the pick procedure applies, we either do random
selection (random pick, \textbf{RP}) or the procedure described in \S\ref{sec:submodular} (submodular pick, \textbf{SP}).
We refer to pick-explainer combinations by adding RP or SP as a prefix. 

\subsection{Are explanations faithful to the model?}
\label{sec:simulated:faithful}
\begin{figure}
\centering
\subfloat[Sparse LR]{
\includegraphics[width=0.45\columnwidth]{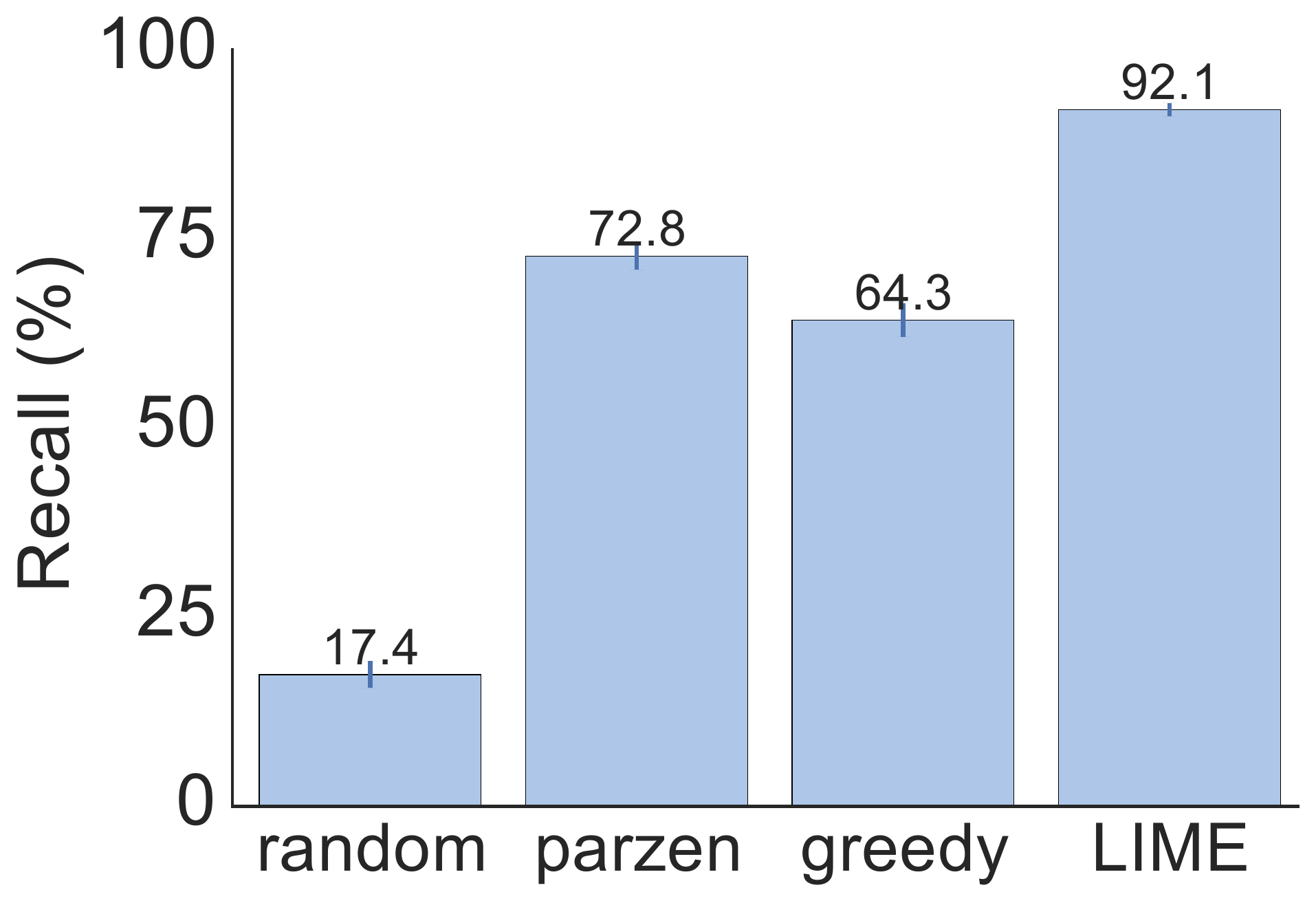}
}\quad
\subfloat[Decision Tree]{
\includegraphics[width=0.45\columnwidth]{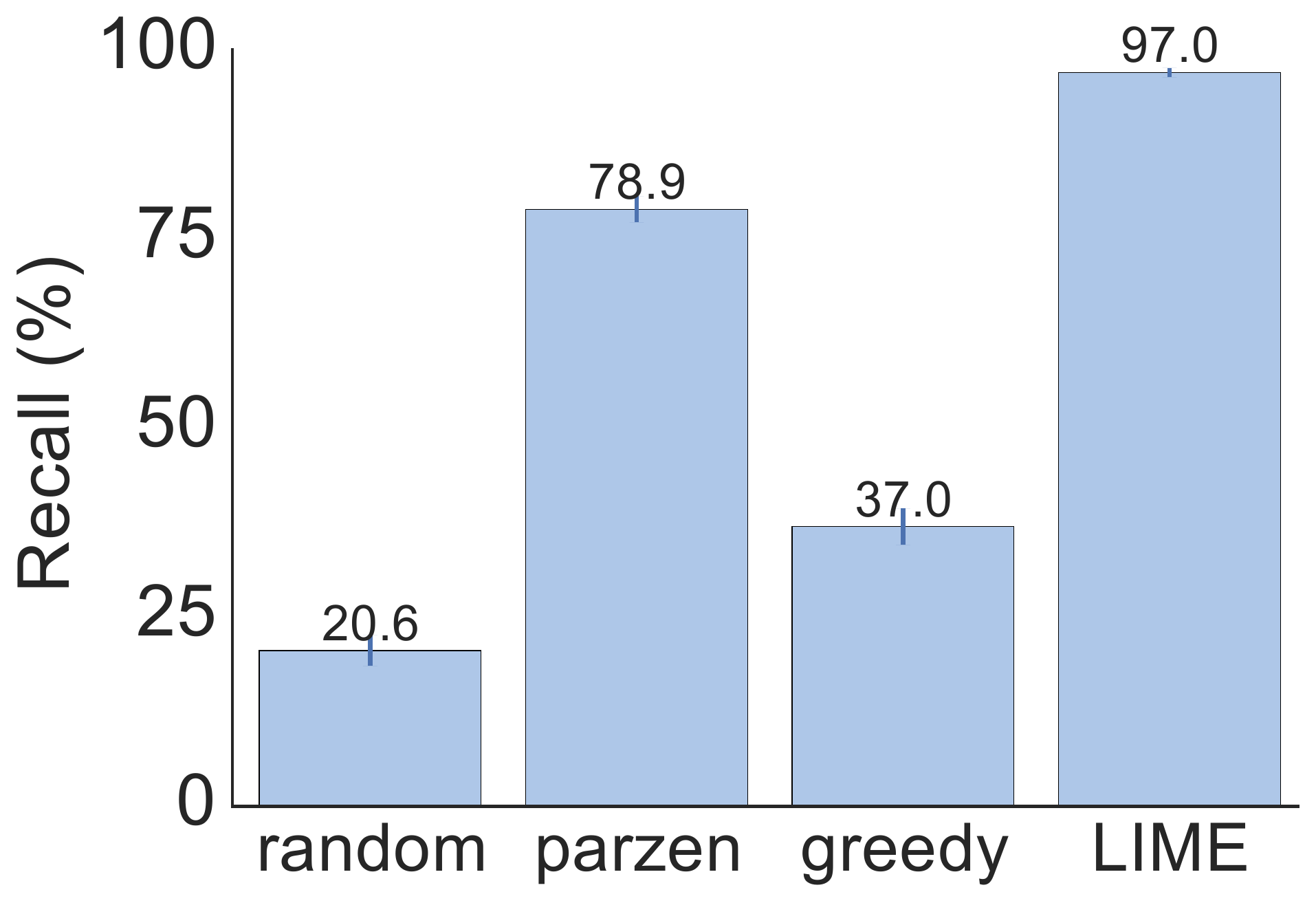}
}
\precap
\caption{Recall on truly important features for two interpretable classifiers on the \textbf{books} dataset. \label{fig:faithfulnessbook}}
\postcap
\end{figure}

\begin{figure}
\centering
\subfloat[Sparse LR]{
\includegraphics[width=0.45\columnwidth]{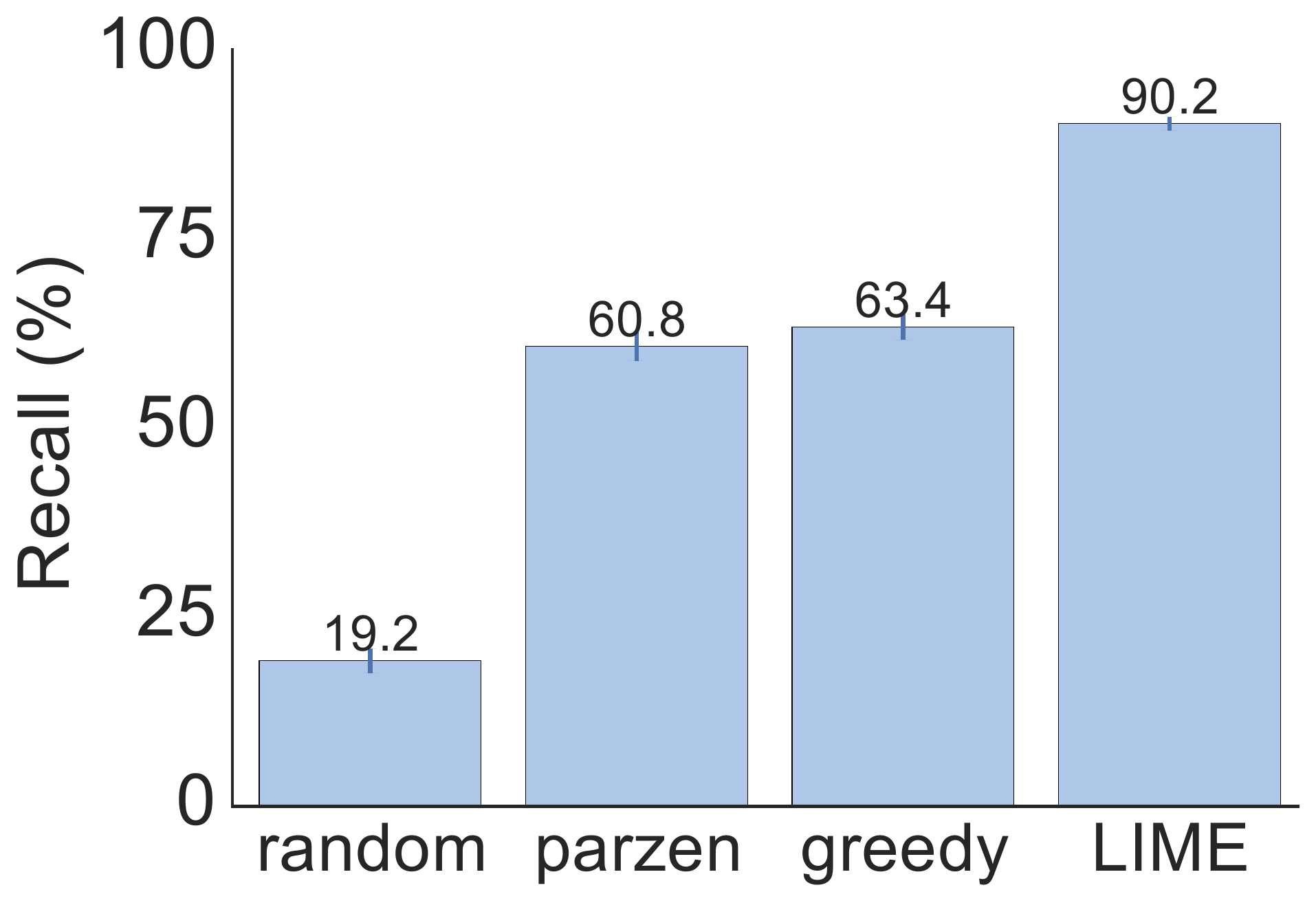}
}\quad
\subfloat[Decision Tree]{
\includegraphics[width=0.45\columnwidth]{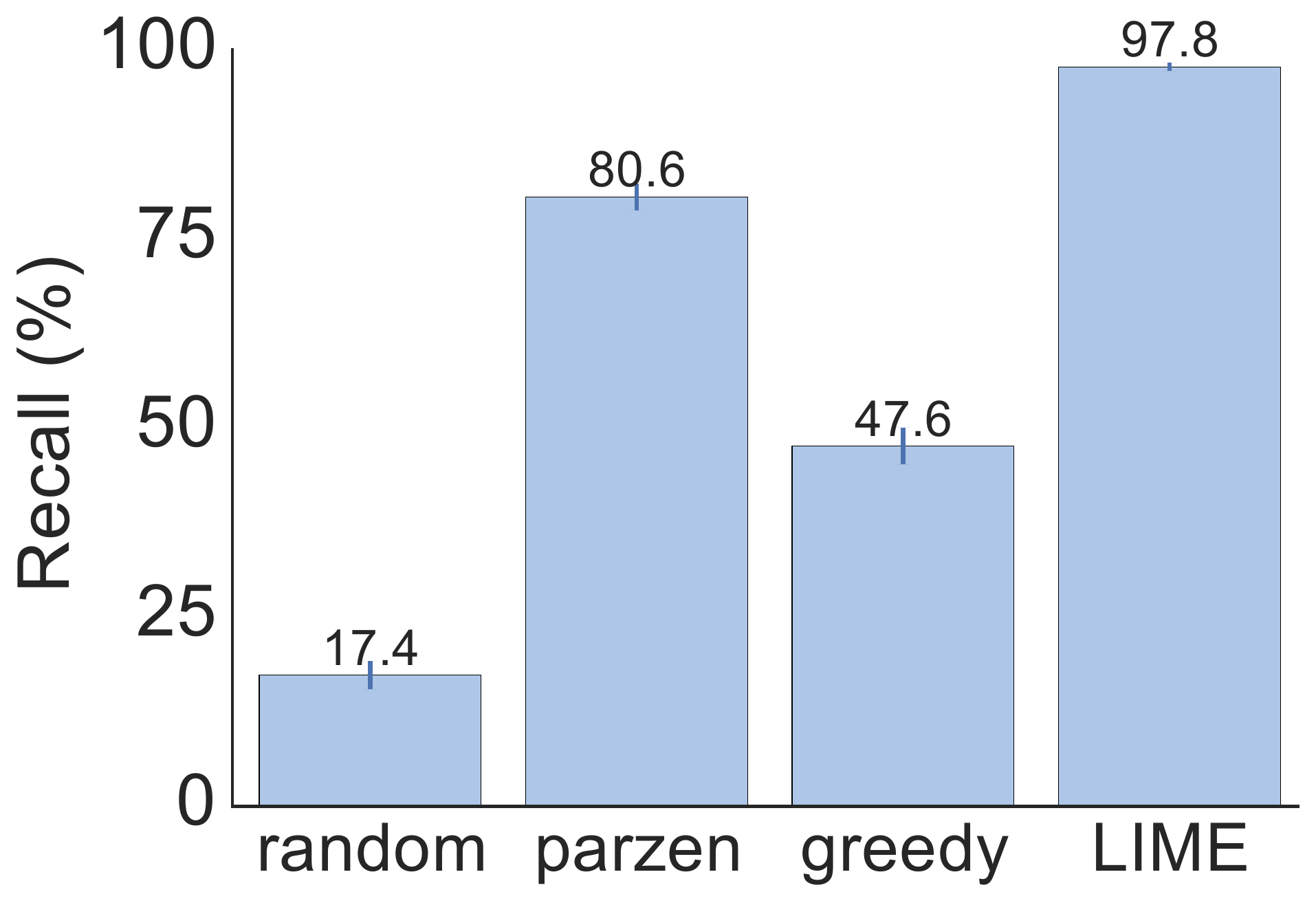}
}
\precap
\caption{Recall on truly important features for two interpretable classifiers on the \textbf{DVDs} dataset.  \label{fig:faithfulnessdvd}}
\postcap
\end{figure}

We measure faithfulness of explanations on classifiers that are by themselves interpretable (sparse logistic regression and decision trees). 
In particular, we train both classifiers such that the maximum number of features they use for any instance is $10$,
and thus we know the \emph{gold} set of features that the are considered important by these models.
For each prediction on the test set, we generate explanations and compute the fraction of these \emph{gold} features that are recovered by the explanations. 
We report this recall averaged over all the test instances in Figures~\ref{fig:faithfulnessbook} and \ref{fig:faithfulnessdvd}. 
We observe that the greedy approach is comparable to parzen on logistic regression, but is substantially worse on decision trees since changing a single feature at a time often does not have an effect on the prediction.
The overall recall by parzen is low, likely due to the difficulty in approximating the original high-dimensional classifier.
\shortName{} consistently provides $>90\%$ recall for both classifiers on both datasets, demonstrating that \shortName{} explanations are faithful to the models.

\subsection{Should I trust this prediction?}
In order to simulate trust in individual predictions, we first randomly select $25\%$ of the features to be ``untrustworthy'', and assume that the users can identify and would not want to trust these features (such as the headers in 20 newsgroups, leaked data, etc).
We thus develop \emph{oracle} ``trustworthiness'' by labeling test set predictions from a black box classifier as ``untrustworthy'' if the prediction changes when untrustworthy features are removed from the instance, and ``trustworthy'' otherwise.
In order to simulate users, we assume that users deem predictions untrustworthy from \shortName{} and parzen explanations if the prediction from the linear approximation changes when all untrustworthy features that appear in the explanations are removed (the simulated human ``discounts'' the effect of untrustworthy features).
For greedy and random, the prediction is mistrusted if any untrustworthy features are present in the explanation, since these methods do not provide a notion of the contribution of each feature to the prediction.
Thus for each test set prediction, we can evaluate whether the simulated user trusts it using each explanation method, and compare it to the trustworthiness oracle.

Using this setup, we report the F1 on the trustworthy predictions for each explanation method, averaged over $100$ runs, in Table \ref{tab:f1}.
The results indicate that \shortName{} dominates others (all results are significant at $p=0.01$) on both datasets, and for all of the black box models.
The other methods either achieve a lower recall (i.e. they mistrust predictions more than they should) or lower precision (i.e. they trust too many predictions), while \shortName{} maintains both high precision and high recall. 
Even though we artificially select which features are untrustworthy, these results indicate that \shortName{} is helpful in assessing trust in individual predictions.

\begin{table}[tb] 
    \centering
    \caption{Average F1 of \emph{trustworthiness} for different explainers on a collection of classifiers and datasets.}~\label{tab:f1}
    {\small
    \begin{tabular}{l
    c@{\hskip0.8\tabcolsep}
    c@{\hskip0.8\tabcolsep}
    c@{\hskip0.8\tabcolsep}
    c
    c@{\hskip0.8\tabcolsep}
    c@{\hskip0.8\tabcolsep}
    c@{\hskip0.8\tabcolsep}
    c}
      \toprule
      & \multicolumn{4}{c}{\bf Books}
      & \multicolumn{4}{c}{\bf DVDs}
      \\
      \cmidrule(lr){2-5} \cmidrule(lr){6-9}
      & {{LR}}
      & {{NN}}
      & {{RF}}
      & {{SVM}}
      & {{LR}}
      & {{NN}}
      & {{RF}}
      & {{SVM}}\\
      \midrule
        Random & 14.6 & 14.8 &14.7 &14.7 & 14.2 & 14.3 &14.5 &14.4\\
        Parzen & 84.0 & 87.6 &94.3 &92.3& 87.0 & 81.7 &94.2 &87.3\\
        Greedy & 53.7 & 47.4 &45.0 &53.3 & 52.4 & 58.1 &46.6 &55.1\\
        LIME & \textbf{96.6} & \bf94.5 &\bf96.2 & \bf 96.7 & \textbf{96.6} & \bf91.8 &\bf96.1 & \bf 95.6\\
        \bottomrule
    \end{tabular}
    }
\end{table}

\subsection{Can I trust this model?}
In the final simulated user experiment, we evaluate whether the explanations can be used for model selection, simulating the case where a human has to decide between two competing models with similar accuracy on validation data. 
For this purpose, we add $10$ artificially ``noisy'' features.
Specifically, on training and validation sets ($80/20$ split of the original training data), each artificial feature appears in $10\%$ of the examples in one class, and $20\%$ of the other, while on the test instances, each artificial feature appears in $10\%$ of the examples in each class. 
This recreates the situation where the models use not only features that are informative in the real world, but also ones that introduce spurious correlations. 
We create pairs of competing classifiers by repeatedly training pairs of random forests with $30$ trees until their validation accuracy is within $0.1\%$ of each other, but their test accuracy differs by at least $5\%$.
Thus, it is not possible to identify the \emph{better} classifier (the one with higher test accuracy) from the accuracy on the validation data.

\begin{figure}[tb]
    \centering
    \subfloat[Books dataset]{
    \includegraphics[width=0.495\columnwidth]{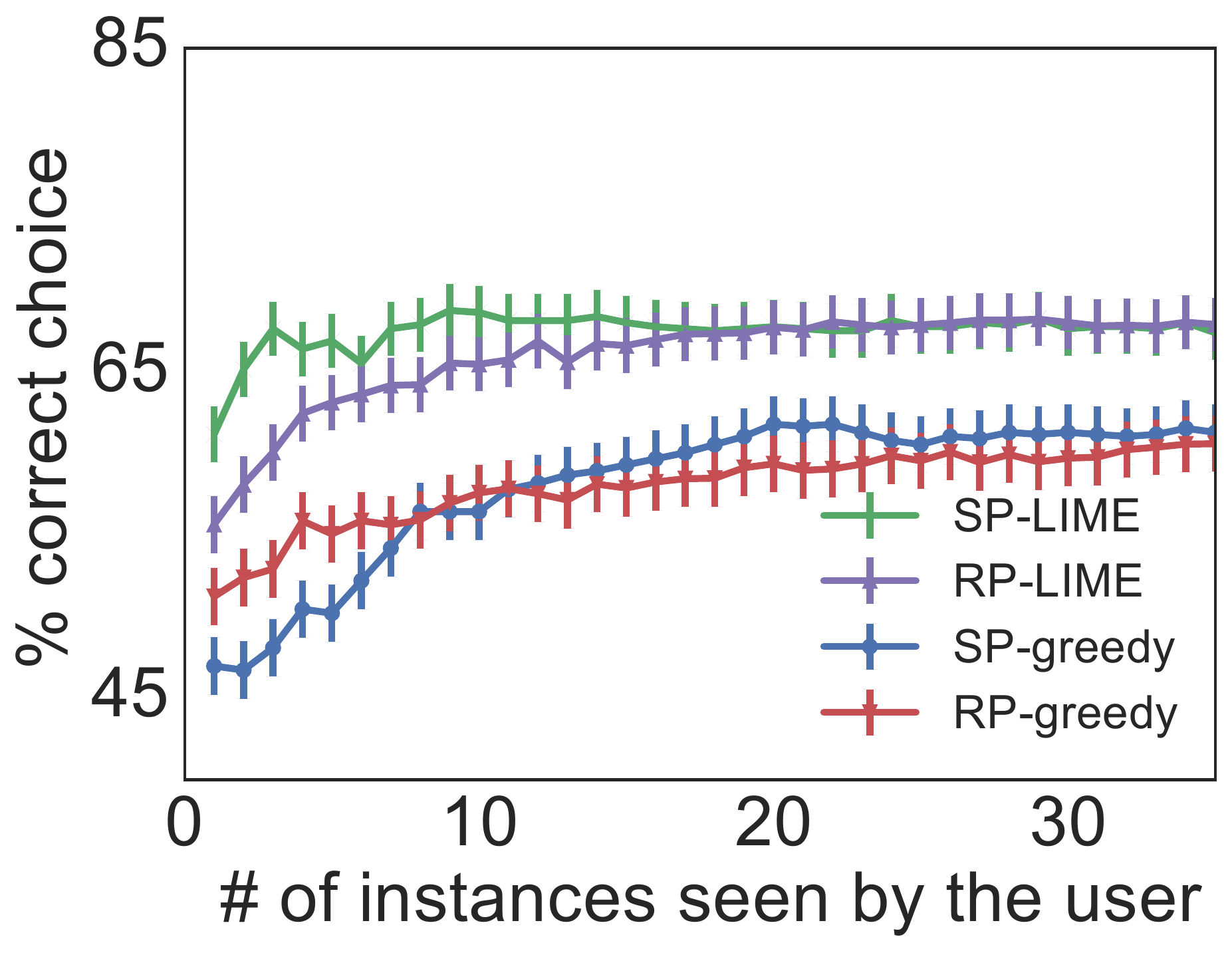}
    }
    \subfloat[DVDs dataset]{
    \includegraphics[width=0.495\columnwidth]{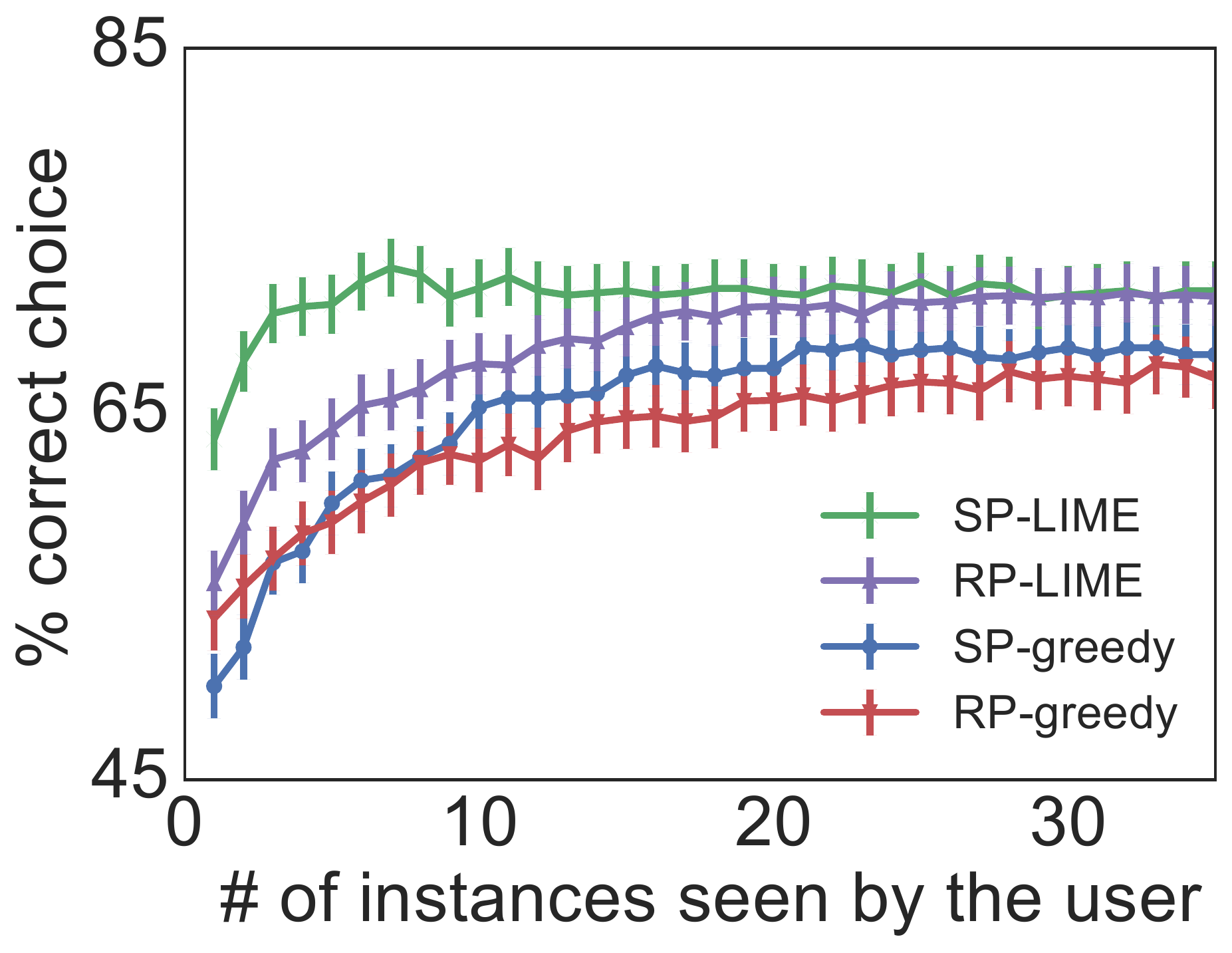}
    }
    \precap
    \caption{Choosing between two classifiers, as the number of instances
    shown to a simulated user is varied. Averages and
    standard errors from 800 runs.}\label{fig:pickn}
    \postcap
\end{figure}

\balance{} 

The goal of this experiment is to evaluate whether a user can identify the better classifier based on the explanations of $\instanceBudget{}$ instances from the validation set.
The simulated human marks the set of artificial features that appear in the $\instanceBudget{}$ explanations as untrustworthy, following which we evaluate how many total predictions in the validation set should be trusted (as in the previous section, treating only marked features as untrustworthy).
Then, we select the classifier with fewer untrustworthy predictions, and compare this choice to the classifier with higher held-out test set accuracy.

We present the accuracy of picking the correct classifier as $\instanceBudget{}$ varies, averaged over $800$ runs, in Figure~\ref{fig:pickn}.
We omit SP-parzen and RP-parzen from the figure since they did not produce useful explanations, performing only slightly better than random. 
\shortName{} is consistently better than greedy, irrespective of the pick method. 
Further, combining submodular pick with \shortName{} outperforms all other methods, in particular it is much better than RP-\shortName{} when only a few examples are shown to the users.
These results demonstrate that the trust assessments provided by SP-selected \shortName{} explanations are good indicators of generalization, which we validate with human experiments in the next section.
\presec
\section{Evaluation with human subjects}
\label{sec:human}

In this section, we recreate three scenarios in machine learning that require trust and understanding of predictions and models.
In particular, we evaluate \shortName{} and SP-\shortName{} in the following settings:
(1)~Can users choose which of two classifiers generalizes better (\S~\ref{sec:human:trustv2}),
(2)~based on the explanations, can users perform feature engineering to improve the model (\S~\ref{sec:human:clean}), and 
(3)~are users able to identify and describe classifier irregularities by looking at explanations (\S~\ref{sec:human:insight}).

\subsection{Experiment setup}
\label{sec:human:setup}
For experiments in \S\ref{sec:human:trustv2} and \S\ref{sec:human:clean}, we use the ``Christianity'' and ``Atheism'' documents from the 20 newsgroups dataset mentioned beforehand. 
This dataset is problematic since it contains features that do not generalize (e.g. very informative header information and author names), and thus validation accuracy considerably overestimates real-world performance. 

In order to estimate the real world performance, we create a new \emph{religion dataset} for evaluation.
We download Atheism and Christianity websites from the DMOZ directory and human curated lists, yielding $819$ webpages in each class.
High accuracy on this dataset by a classifier trained on 20 newsgroups indicates that the classifier is generalizing using semantic content, instead of placing importance on the data specific issues outlined above.
Unless noted otherwise, we use {SVM} with RBF kernel, trained on the 20 newsgroups data with hyper-parameters tuned via the cross-validation.

\subsection{Can users select the best classifier?}
\label{sec:human:trustv2}

In this section, we want to evaluate whether explanations can help users decide which classifier generalizes better, i.e., which classifier would the user deploy ``in the wild''.
Specifically, users have to decide between two classifiers: {SVM} trained on the original 20 newsgroups dataset, and a version of the same classifier trained on a ``cleaned'' dataset where many of the features that do not generalize have been manually removed. 
The original classifier achieves an accuracy score of $57.3\%$ on the \emph{religion dataset}, while the ``cleaned'' classifier achieves a score of $69.0\%$.
In contrast, the test accuracy on the original 20 newsgroups split is $94.0\%$ and $88.6\%$, respectively -- suggesting that the worse classifier would be selected if accuracy alone is used as a measure of trust. 

We recruit human subjects on Amazon Mechanical Turk -- by no means machine learning experts, but instead people with basic knowledge about religion. 
We measure their ability to choose the better algorithm by seeing side-by-side explanations with the associated raw data (as shown in Figure~\ref{fig:mturkinterface1}). 
We restrict both the number of words in each explanation~($K$) and the number of documents that each person inspects~($\instanceBudget$) to $6$.
The position of each algorithm and the order of the instances seen are randomized between subjects.
After examining the explanations, users are asked to select which algorithm will perform best in the real world. 
The explanations are produced by either {greedy} (chosen as a baseline due to its performance in the simulated user experiment) or {\shortName{}}, and the instances are selected either by random (RP) or submodular pick (SP).
We modify the greedy step in Algorithm~\ref{alg:submod} slightly so it alternates between explanations of the two classifiers.  For each setting, we repeat the experiment with $100$ users.

\begin{figure}[tb]
    \centering
    \includegraphics[width=0.3\textwidth]{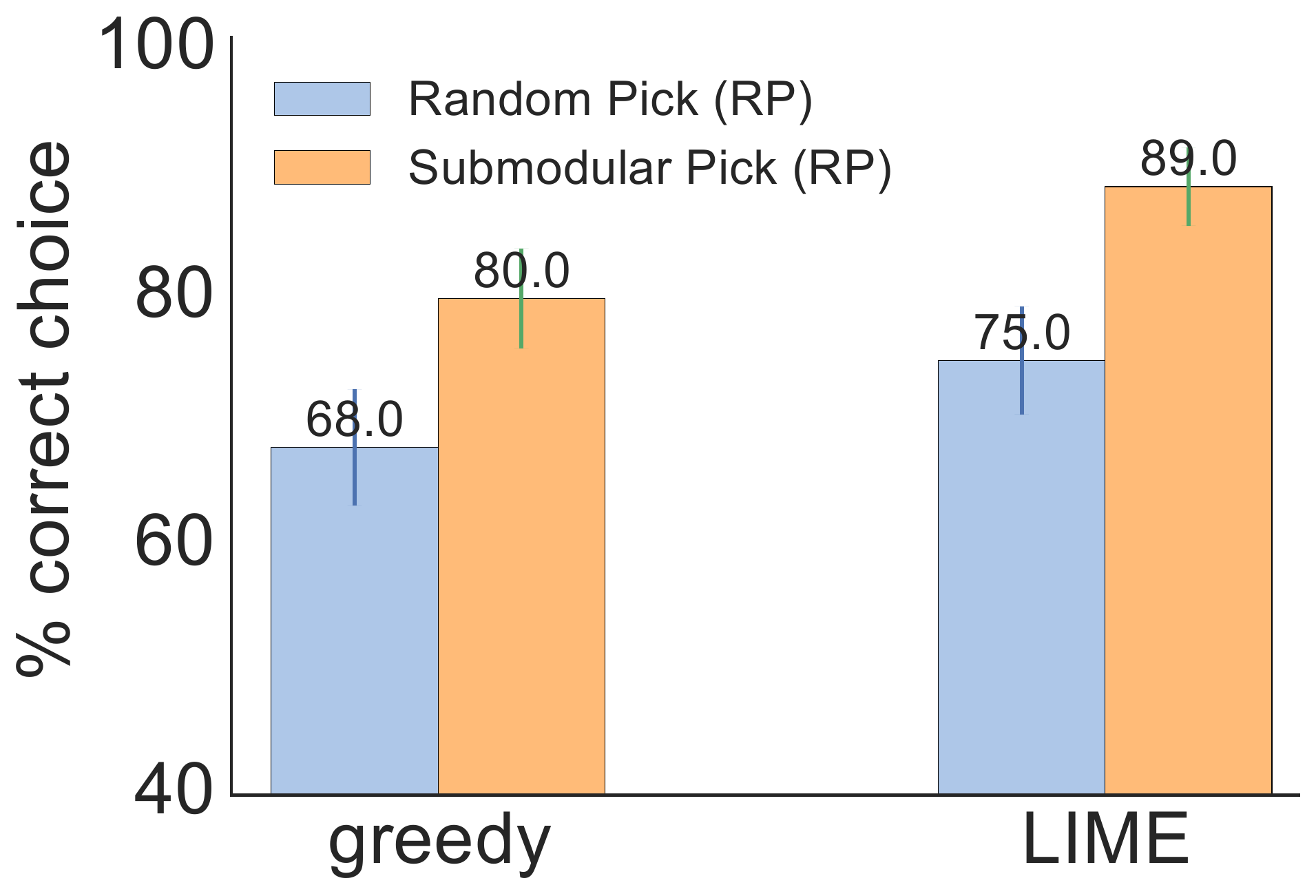}
    \precap
    \caption{Average accuracy of human subject (with standard errors) in choosing between two classifiers.}
    \postcapsmall
    
    \label{fig:mturk1_results}
\end{figure}

The results are presented in Figure~\ref{fig:mturk1_results}.
Note that all of the methods are good at identifying the better classifier, demonstrating that the explanations are useful in determining which classifier to trust, while using test set accuracy would result in the selection of the wrong classifier.
Further, we see that the submodular pick~(SP) greatly improves the user's ability to select the best classifier when compared to random pick~(RP), with \shortName{} outperforming {greedy} in both cases.

\begin{figure}[tb]
    \centering
    \includegraphics[width=.4\textwidth]{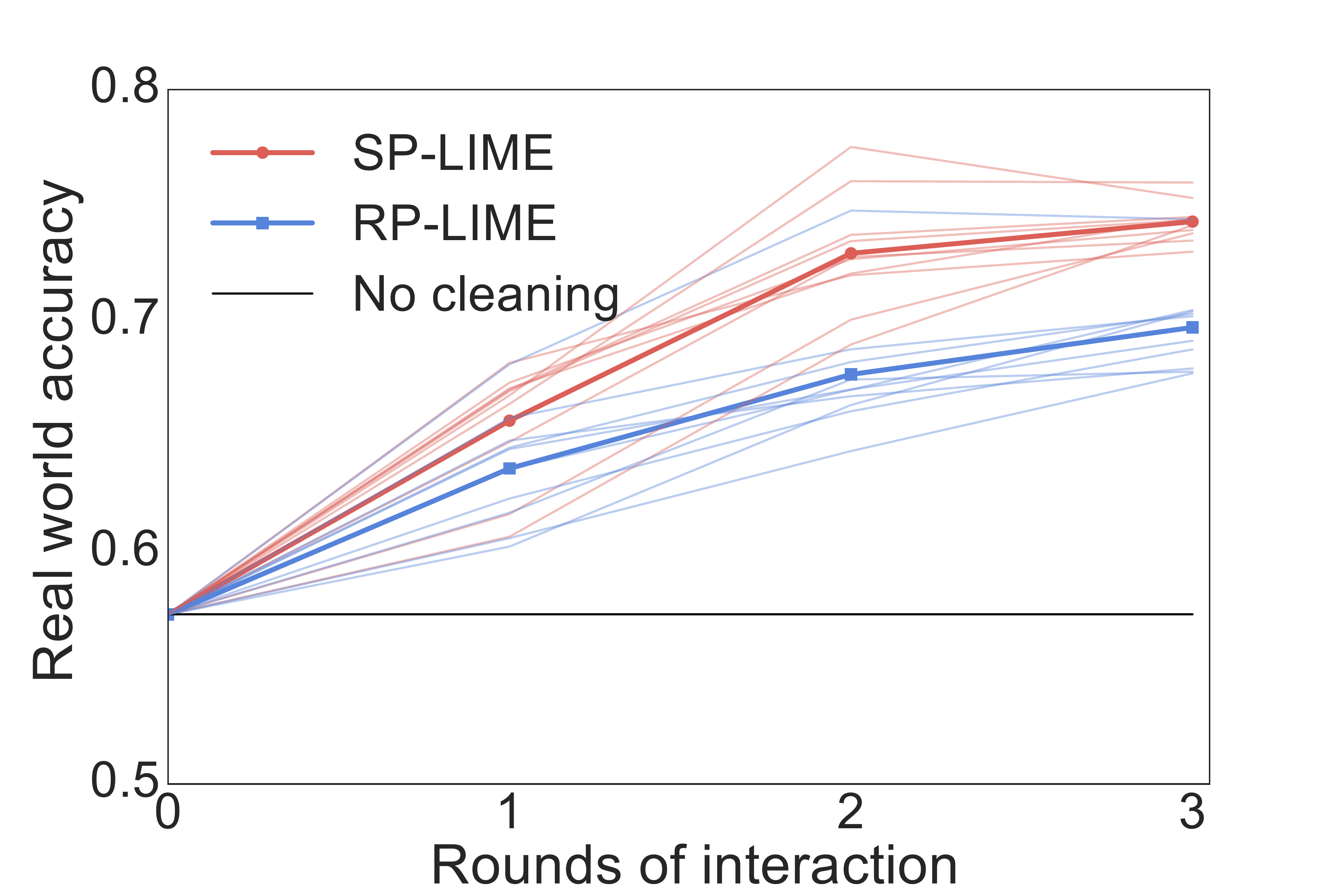}
    \precap
    \caption{Feature engineering experiment. Each shaded line represents the average accuracy of subjects in a path starting from one of the initial $10$ subjects. Each solid line represents the average across all paths per round of interaction.}~\label{fig:mturk2_result}
    \postcap
\end{figure}

\subsection{Can non-experts improve a classifier?}
\label{sec:human:clean}
If one notes that a classifier is untrustworthy, a common task in machine learning is feature engineering, i.e. modifying the set of features and retraining in order to improve generalization. 
Explanations can aid in this process by presenting the important features, particularly for removing features that the users feel do not generalize.

We use the 20 newsgroups data here as well, and ask Amazon Mechanical Turk users to identify which words from the explanations should be removed from subsequent training, for the worse classifier from the previous section (\S\ref{sec:human:trustv2}).
In each round, the subject marks words for deletion after observing $\instanceBudget=10$ instances with $K=10$ words in each explanation (an interface similar to Figure~\ref{fig:mturkinterface1}, but with a single algorithm). 
As a reminder, the users here are not experts in machine learning and are unfamiliar with feature engineering, thus are only identifying words based on their semantic content.
Further, users do not have any access to the \emph{religion} dataset -- they do not even know of its existence.
We start the experiment with $10$ subjects.
After they mark words for deletion, we train $10$ different classifiers, one for each subject (with the corresponding words removed).
The explanations for each classifier are then presented to a set of $5$ users in a new round of interaction, which results in $50$ new classifiers.
We do a final round, after which we have $250$ classifiers, each with a path of interaction tracing back to the first $10$ subjects.

The explanations and instances shown to each user are produced by \textbf{SP-LIME} or \textbf{RP-LIME}.
We show the average accuracy on the \emph{religion} dataset at each interaction round for the paths originating from each of the original $10$ subjects (shaded lines), and the average across all paths (solid lines) in Figure \ref{fig:mturk2_result}.
It is clear from the figure that the crowd workers are able to improve the model by removing features they deem unimportant for the task.
Further, \textbf{SP-LIME} outperforms \textbf{RP-LIME}, indicating selection of the instances to show the users is crucial for efficient feature engineering.

Each subject took an average of $3.6$ minutes per round of cleaning, resulting in just under 11 minutes to produce a classifier that generalizes much better to real world data. 
Each path had on average $200$ words removed with \textbf{SP}, and $157$ with
\textbf{RP}, indicating that incorporating coverage of important features is useful for feature engineering.
Further, out of an average of $200$ words selected with SP, $174$ were selected by at least half of the users, while $68$ by \emph{all} the users.
Along with the fact that the variance in the accuracy decreases across rounds, this high agreement demonstrates that the users are converging to similar \emph{correct} models.
This evaluation is an example of how explanations make it easy to improve an untrustworthy classifier -- in this case easy enough that machine learning knowledge is not required.

\subsection{Do explanations lead to insights?}
\label{sec:human:insight}

Often artifacts of data collection can induce undesirable correlations that the classifiers pick up during training.
These issues can be very difficult to identify just by looking at the raw data and predictions.
In an effort to reproduce such a setting, we take the task of distinguishing between photos of Wolves and Eskimo Dogs (huskies).
We train a logistic regression classifier on a training set of $20$ images, hand selected such that all pictures of wolves had snow in the background, while pictures of huskies did not.
As the features for the images, we use the first max-pooling layer of Google's pre-trained Inception neural network~\cite{inception}.
On a collection of additional $60$ images, the classifier predicts ``Wolf'' if there is snow (or light background at the bottom), and ``Husky'' otherwise, regardless of animal color, position, pose, etc.
We trained this \emph{bad} classifier intentionally, to evaluate whether subjects are able to detect it.

\begin{figure}
    \centering
    \subfloat[Husky classified as wolf]{
    \includegraphics[width=0.42\columnwidth]{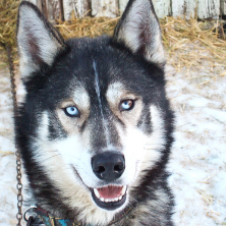}
    \label{fig:husky1}
    }\quad
    \subfloat[Explanation]{
    \includegraphics[width=0.42\columnwidth]{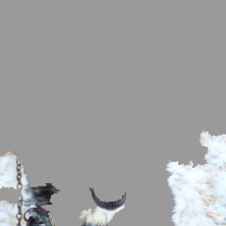}
    \label{fig:exphusky}
    }\\

    \precap
    \caption{Raw data and explanation of a bad model's prediction in the ``Husky vs Wolf'' task.}
\end{figure}

The experiment proceeds as follows: we first present a balanced set of $10$ test predictions (without explanations), where one wolf is not in a snowy background (and thus the prediction is ``Husky'') and one husky is (and is thus predicted as ``Wolf'').
We show the ``Husky'' mistake in Figure \ref{fig:husky1}.
The other $8$ examples are classified correctly.
We then ask the subject three questions: (1) Do they trust this algorithm to work well in the real world,
(2) why,
and (3) how do they think the algorithm is able to distinguish between these photos of wolves and huskies. 
After getting these responses, we show the same images with the associated explanations, such as in Figure \ref{fig:exphusky}, and ask the same questions.

Since this task requires some familiarity with the notion of spurious correlations and generalization, the set of subjects for this experiment were graduate students who have taken at least one graduate machine learning course. 
After gathering the responses, we had $3$ independent evaluators read their reasoning and determine if each subject mentioned snow, background, or equivalent as a feature the model may be using. 
We pick the majority to decide whether the subject was correct about the insight, and report these numbers before and after showing the explanations in Table \ref{tab:mturk3}. 

\begin{table}[tb]
\begin{tabular}{lcc}
\toprule
&Before & After \\ 
\midrule
Trusted the bad model & 10 out of 27 & 3 out of 27 \\ 
Snow as a potential feature &12 out of 27 & 25 out of 27\\
\bottomrule
\end{tabular}
\caption{``Husky vs Wolf'' experiment results.}
\label{tab:mturk3}
\end{table}

Before observing the explanations, more than a third trusted the classifier, 
and a little less than half mentioned the snow pattern as something the neural network was using -- although all speculated on other patterns. 
After examining the explanations, however, almost all of the subjects identified the correct insight, with much more certainty that it was a determining factor.
Further, the trust in the classifier also dropped substantially.
Although our sample size is small, this experiment demonstrates the utility of explaining individual predictions for getting insights into classifiers knowing when not to trust them and why. 
\section{Related Work}
\label{sec:related}
The problems with relying on validation set accuracy as the primary measure of trust have been well studied. 
Practitioners consistently overestimate their model's accuracy~\cite{Patel:2008:ISM:1357054.1357160}, propagate feedback loops~\cite{technical_debt}, or fail to notice data leaks~\cite{leakage}. 
In order to address these issues, researchers have proposed tools like Gestalt~\cite{gestalt} and Modeltracker~\cite{modeltracker}, which help users navigate individual instances. 
These tools are complementary to {\shortName{}} in terms of explaining models, since they do not address the problem of explaining individual predictions. 
Further, our submodular pick procedure can be incorporated in such tools to aid users in navigating larger datasets.

Some recent work aims to anticipate failures in machine learning, specifically for vision tasks~\cite{bansal2014transparent, failuresvision}. 
Letting users know when the systems are likely to fail can lead to an increase in trust, by avoiding ``silly mistakes''~\cite{dzindolet}.
These solutions either require additional annotations and feature engineering that is specific to vision tasks or do not provide insight into why a decision should not be trusted.
Furthermore, they assume that the current evaluation metrics are reliable, which may not be the case if problems such as data leakage are present.
Other recent work~\cite{pick_kulesza} focuses on exposing users to different kinds of mistakes (our pick step).
Interestingly, the subjects in their study did not notice the serious problems in the 20 newsgroups data even after looking at many mistakes, suggesting that examining raw data is not sufficient.
Note that \citet{pick_kulesza} are not alone in this regard, many researchers in the field have unwittingly published classifiers that would not generalize for this task. 
Using \shortName{}, we show that even non-experts are able to identify these irregularities when explanations are present.
Further, {\shortName{}} can complement these existing systems, and allow users to assess trust even when a prediction seems ``correct'' but is made for the wrong reasons.

Recognizing the utility of explanations in assessing trust, many have proposed using interpretable models~\cite{WangRu15}, especially for the medical domain~\cite{caruana2015, LethamRuMcMa15,supersparse}.
While such models may be appropriate for some domains, they may not apply equally well to others (e.g. a supersparse linear model \cite{supersparse} with $5-10$ features is unsuitable for text applications).
Interpretability, in these cases, comes at the cost of flexibility, accuracy, or efficiency.
For text, EluciDebug~\cite{EluciDebug} is a full human-in-the-loop system that shares many of our goals (interpretability, faithfulness, etc).
However, they focus on an already interpretable model (Naive Bayes).
In computer vision, systems that rely on object detection to produce candidate alignments~\cite{deepvisual} or attention~\cite{Xu2015show} are able to produce explanations for their predictions.
These are, however, constrained to specific neural network architectures or incapable of detecting ``non object'' parts of the images. 
Here we focus on general, model-agnostic explanations that can be applied to any classifier or regressor that is appropriate for the domain - even ones that are yet to be proposed.

A common approach to model-agnostic explanation is learning a potentially interpretable model on the predictions of the original model~\cite{Baehrens:2010:EIC:1756006.1859912,craven,explain:krr15}.
Having the explanation be a gradient vector~\cite{Baehrens:2010:EIC:1756006.1859912} captures a similar locality intuition to that of {\shortName{}}. 
However, interpreting the coefficients on the gradient is difficult, particularly for confident predictions (where gradient is near zero).
Further, these explanations approximate the original model \emph{globally}, thus maintaining local fidelity becomes a significant challenge, as our experiments demonstrate.
In contrast, {\shortName{}} solves the much more feasible task of finding a model that approximates the original model \emph{locally}.
The idea of perturbing inputs for explanations has been explored before~\cite{gametheory}, where the authors focus on learning a specific \emph{contribution} model, as opposed to our general framework.  
None of these approaches explicitly take cognitive limitations into account, and thus may produce non-interpretable explanations, such as a gradients or linear models with thousands of non-zero weights.
The problem becomes worse if the original features are nonsensical to humans (e.g. word embeddings).
In contrast, {\shortName{}} incorporates interpretability both in the optimization and in our notion of \emph{interpretable representation}, such that domain and task specific interpretability criteria can be accommodated.

\section{Conclusion and Future Work}
\label{sec:conclusions}
\noindent
In this paper, we argued that trust is crucial for effective human interaction with machine learning systems, and that explaining individual predictions is important in assessing trust.
We proposed {\shortName{}}, a modular and extensible approach to faithfully explain the predictions of \emph{any} model in an interpretable manner.
We also introduced SP-\shortName{}, a method to select representative and non-redundant predictions, providing a global view of the model to users.
Our experiments demonstrated that explanations are useful for a variety of models in trust-related tasks in the text and image domains, with both expert and non-expert users: deciding between models, assessing trust, improving untrustworthy models, and getting insights into predictions.

There are a number of avenues of future work that we would like to explore.
Although we describe only sparse linear models as explanations, our framework supports the exploration of a variety of explanation families, such as decision trees; it would be interesting to see a comparative study on these with real users.
One issue that we do not mention in this work was how to perform the pick step for images, and we would like to address this limitation in the future.
The domain and model agnosticism enables us to explore a variety of applications, and we would like to investigate potential uses in speech, video, and medical domains, as well as recommendation systems.
Finally, we would like to explore theoretical properties (such as the appropriate number of samples) and computational optimizations (such as using parallelization and GPU processing), in order to provide the accurate, real-time explanations that are critical for any human-in-the-loop machine learning system.

\vspace{-0.05in}
\section*{Acknowledgements}

We would like to thank Scott Lundberg, Tianqi Chen, and Tyler Johnson for helpful discussions and feedback.
This work was supported in part by ONR awards \#W911NF-13-1-0246 and \#N00014-13-1-0023, and in part by TerraSwarm, one of six centers of STARnet, a Semiconductor Research Corporation program sponsored by MARCO and DARPA.

\small
\bibliographystyle{abbrvnat}
\bibliography{sample}  
\end{document}